\documentclass[10pt,twocolumn,letterpaper]{article}

\usepackage{cvpr}
\usepackage{times}
\usepackage{epsfig}
\usepackage{graphicx}
\usepackage{amsmath}
\usepackage{amssymb}
\usepackage{bm}
\usepackage{booktabs}
\usepackage{siunitx}
\usepackage{stfloats}
\usepackage{multirow}

\DeclareMathOperator*{\argmax}{argmax} 

\usepackage[breaklinks=true,bookmarks=false]{hyperref}

\cvprfinalcopy 

\def\cvprPaperID{****} 

\ifcvprfinal\fi
\begin{document}

\title{Dynamic Convolution: Attention over Convolution Kernels}

\author{{Yinpeng Chen \qquad Xiyang Dai \qquad Mengchen Liu \qquad Dongdong Chen \qquad Lu Yuan \qquad Zicheng Liu}\\
Microsoft\\
{\tt\small \{yiche,xidai,mengcliu,dochen,luyuan,zliu\}@microsoft.com}
}

\maketitle

\begin{abstract}
Light-weight convolutional neural networks (CNNs) suffer performance degradation as their low computational budgets constrain both the depth (number of convolution layers) and the width (number of channels) of CNNs, resulting in limited representation capability. To address this issue, we present Dynamic Convolution, a new design that increases model complexity without increasing the network depth or width. Instead of using a single convolution kernel per layer, dynamic convolution aggregates multiple parallel convolution kernels dynamically based upon their attentions, which are input dependent. Assembling multiple kernels is not only computationally efficient due to the small kernel size, but also has more representation power since these kernels are aggregated in a non-linear way via attention. By simply using dynamic convolution for the state-of-the-art architecture MobileNetV3-Small, the top-1 accuracy of ImageNet classification is boosted by 2.9\% with only 4\% additional FLOPs and 2.9 AP gain is achieved on COCO keypoint detection.
\end{abstract}
\section{Introduction}
Interest in building light-weight and efficient neural networks has exploded recently. It not only enables new experiences on mobile devices, but also protects user's privacy from sending personal information to the cloud. Recent works (e.g. MobileNet \cite{howard2017mobilenets, sandler2018mobilenetv2, howard2019mbnetv3} and ShuffleNet \cite{Zhang_2018_CVPR, Ma_2018_ECCV}) have shown that both efficient operator design (e.g. depthwise convolution, channel shuffle, squeeze-and-excitation \cite{Hu_2018_CVPR}, asymmetric convolution \cite{Ding_2019_ICCV}) and architecture search (\cite{Tan_2019_CVPR, guo2019single, cai2018proxylessnas}) are important for designing efficient convolutional neural networks. 

However, even the state-of-the-art efficient CNNs (e.g. MobileNetV3 \cite{howard2019mbnetv3}) suffer significant performance degradation when the computational constraint becomes extremely low. For instance, when the computational cost of MobileNetV3 reduces from 219M to 66M  Multi-Adds, the top-1 accuracy of ImageNet classification drops from 75.2\% to 67.4\%. This is because the extremely low computational cost severely constrains both the network depth (number of layers) and width (number of channels), which are crucial for the network performance but proportional to the computational cost.  

\begin{figure}[t]
	\begin{center}
		\includegraphics[width=0.8\linewidth]{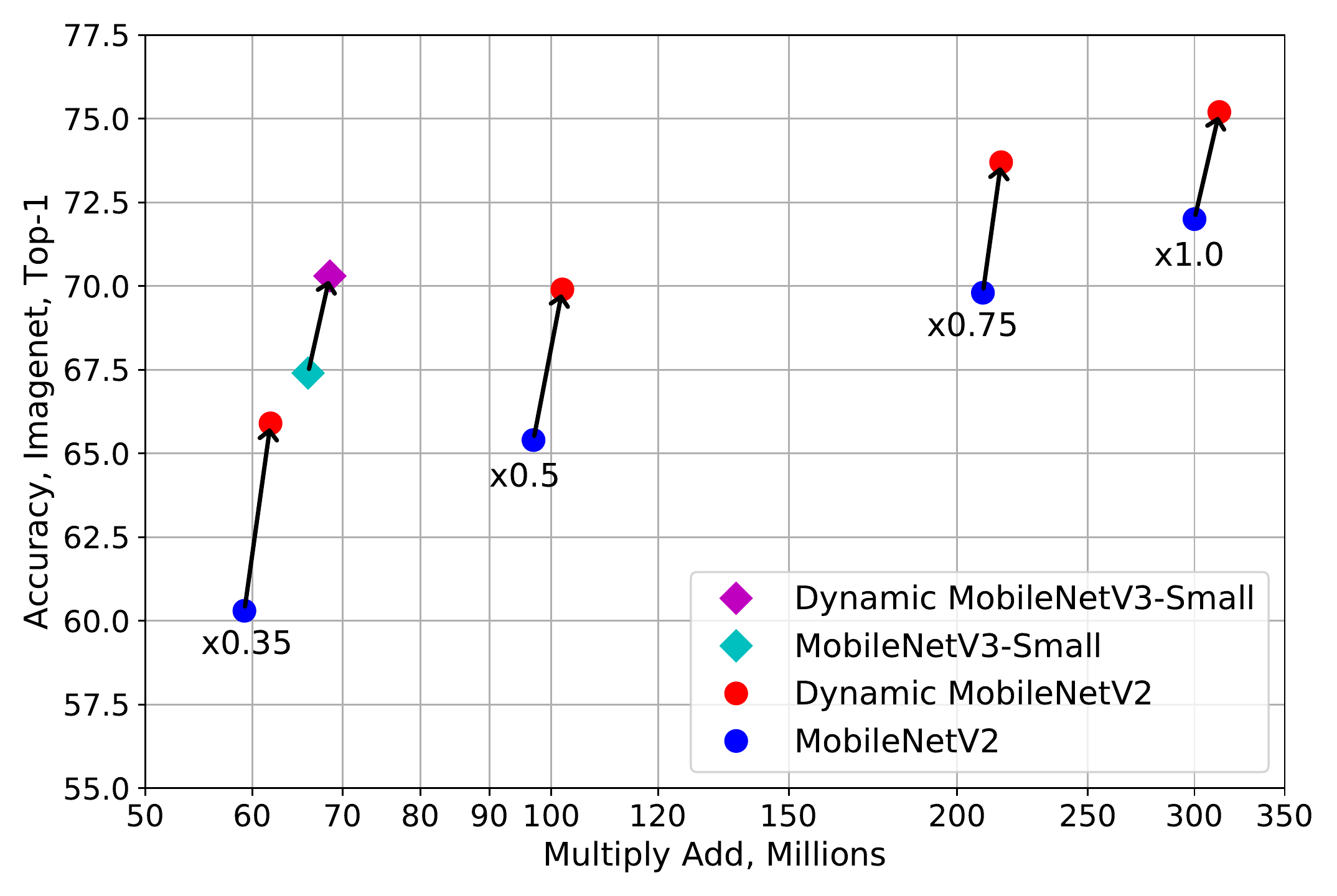}
	\end{center}
	\caption{The trade-off between computational cost (MAdds) and top-1 accuracy of ImageNet classification. Dynamic convolution significantly boosts the accuracy with a small amount of extra MAdds on MobileNet V2 and V3. Best viewed in color.}
	\label{fig:highlight}
\end{figure}

\begin{figure}[t]
	\begin{center}
		\includegraphics[width=0.95\linewidth]{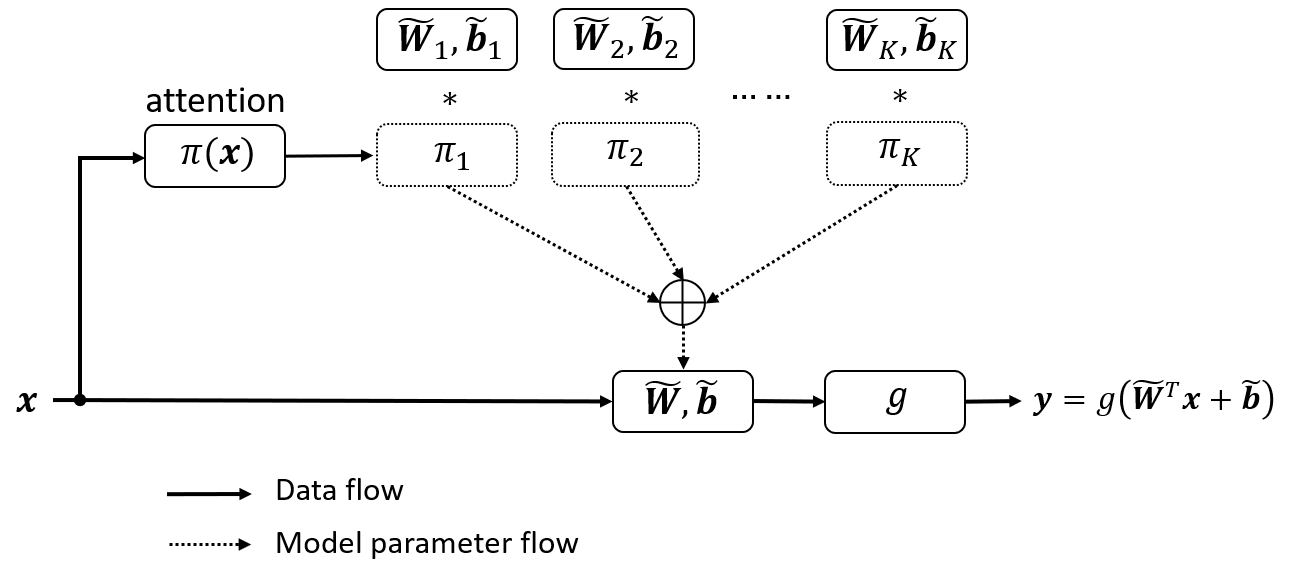}
	\end{center}
	\caption{Dynamic perceptron. It aggregates multiple linear functions dynamically based upon their attentions $\{\pi_k\}$, which are input dependent.}
	\label{fig:overview}
\end{figure}

This paper proposes a new operator design, named \textit{dynamic convolution}, to increase the representation capability with negligible extra FLOPs. Dynamic convolution uses a set of $K$ parallel convolution kernels 
$\{\bm{\tilde{W}}_k, \bm{\tilde{b}}_k\}$ instead of using a single convolution kernel per layer (see Figure \ref{fig:overview}). 
%
These convolution kernels are aggregated dynamically $\bm{\tilde{W}}=\sum_k\pi_k(\bm{x})\bm{\tilde{W}}_k$ for each individual input $\bm{x}$ (e.g. image) via input dependent attention $\pi_k(\bm{x})$. The biases are aggregated using the same attention $\bm{\tilde{b}}=\sum_k\pi_k(\bm{x})\bm{\tilde{b}}_k$. 
%
Dynamic convolution is a non-linear function with more representation power than its static counterpart. Meanwhile, dynamic convolution is computationally efficient. It does not increase the depth or width of the network, as the parallel convolution kernels share the output channels by aggregation. It only introduces extra computational cost to compute attentions $\{\pi_k(\bm{x})\}$ and aggregate kernels 
, which is negligible compared to convolution. \textit{The key insight is that within reasonable cost of model size (as convolution kernels are small), dynamic kernel aggregation provides an efficient way (low extra FLOPs) to boost representation capability.}


Dynamic convolutional neural networks (denoted as DY-CNNs) are more difficult to train, as they require joint optimization of all convolution kernels and the attention across multiple layers. We found two keys for efficient joint optimization: (a) constraining the attention output as $\sum_k \pi_k(\bm{x}) = 1$ to facilitate the learning of attention model $\pi_k(\bm{x})$, and (b) flattening attention (near-uniform) in early training epochs to facilitate the learning of convolution kernels $\{\bm{\tilde{W}}_k, \bm{\tilde{b}}_k\}$. We simply integrate these two keys by using softmax with a large temperature for kernel attention.

We demonstrate the effectiveness of dynamic convolution on both image classification (ImageNet) and keypoint detection (COCO). Without bells and whistles, simply replacing static convolution with dynamic convolution in MobileNet V2 and V3 achieves solid improvement with only a slight increase (4\%) of computational cost (see Figure \ref{fig:highlight}). For instance, with 100M Multi-Adds budget, our method gains 4.5\% and 2.9\% top-1 accuracy on image classification for MobileNetV2 and MobileNetV3, respectively.

\section{Related Work}
\noindent \textbf{Efficient CNNs:} Recently, designing efficient CNN architectures \cite{squeezenet16, howard2017mobilenets, sandler2018mobilenetv2, howard2019mbnetv3, Zhang_2018_CVPR, Ma_2018_ECCV} has been an active research area. SqueezeNet \cite{squeezenet16} reduces the number of parameters by using $1\times1$ convolution extensively in the fire module. MobileNetV1 \cite{howard2017mobilenets} substantially reduces FLOPs by decomposing a $3\times3$ convolution into a depthwise convolution and a pointwise convolution. Based upon this, MobileNetV2 \cite{sandler2018mobilenetv2} introduces inverted residuals and linear bottlenecks. MobileNetV3 \cite{howard2019mbnetv3} applies squeeze-and-excitation \cite{Hu_2018_CVPR} in the residual layer and employs a platform-aware neural architecture approach \cite{Tan_2019_CVPR} to find the optimal network structures. ShuffleNet further reduces MAdds for $1\times1$ convolution by channel shuffle operations. ShiftNet \cite{shift} replaces expensive spatial convolution by the shift operation and pointwise convolutions. 
Compared with existing methods, our dynamic convolution can be used to replace any static convolution kernels (e.g. $1\times1$, $3\times3$, depthwise convolution, group convolution) and is complementary to other advanced operators like squeeze-and-excitation.

\noindent \textbf{Model Compression and Quantization:} Model compression \cite{Han2016deepcompression, Liu_2017_ICCV, He_2018_ECCV} and quantization \cite{NIPS2015BinaryConnect, zhu2017trained, Zhang_2018_ECCV, Yang_2019_CVPR, Wang_2019_CVPR} approaches are also important for learning efficient neural networks. They are complementary to our work, helping reduce the model size for our dynamic convolution method. 

\noindent \textbf{Dynamic Deep Neural Networks:} Our method is related to recent works of dynamic neural networks \cite{NIPS2017_6813, liu2018ddnn, Wang_2018_ECCV, Wu_2018_CVPR, yu2018slimmable, huang2018multiscale} that focus on skipping part of an existing model based on input image. D$^2$NN \cite{liu2018ddnn}, SkipNet \cite{Wang_2018_ECCV} and  BlockDrop \cite{Wu_2018_CVPR} learn an additional controller for skipping decision by using reinforcement learning. MSDNet \cite{huang2018multiscale} allows early-exit based on the current prediction confidence. Slimmable Nets \cite{yu2018slimmable} learns a single neural network executable at different width. Once-for-all \cite{Cai2019OnceFA} proposes a progressive shrinking algorithm to train one network that supports multiple sub-networks. The accuracy for these sub-networks is the same as independently trained networks.
Compared with these works, our method has two major differences. Firstly, our method has dynamic convolution kernels but static network structure, while existing works have static convolution kernels but dynamic network structure. Secondly, our method does \textit{not} require an additional controller. The attention is embedded in each layer, enabling end-to-end training. Compared to the concurrent work \cite{Yang2019CondConvCP}, our method is more efficient with better performance.

\noindent \textbf{Neural Architecture Search:} Recent research works in neural architecture search (NAS) are powerful on finding high-accuracy neural network architectures \cite{Zoph2017NeuralAS, real2018aaai, Zoph_2018_CVPR, liu2018darts, xie2018snas} as well as hardware-aware efficient network architectures \cite{cai2018proxylessnas, Tan_2019_CVPR, Wu_2019_CVPR}. The hardware-aware NAS methods incorporate hardware latency into the architecture search process, by making it differentiable. \cite{guo2019single} proposed single path supernet to optimize all architectures in the search space simultaneously, and then perform evolutionary architecture search to handle computational constraints. Based upon NAS, MobileNetV3 \cite{howard2019mbnetv3} shows significant improvements over human-designed baselines (e.g. MobileNetV2 \cite{sandler2018mobilenetv2}).
Our dynamic convolution method can be easily used in advanced architectures found by NAS. Later in this paper, we will show that dynamic convolution not only improves the performance for human-designed networks (e.g. MobielNetV2), but also boosts the performance for automatically searched architectures (e.g. MobileNetV3), with low extra FLOPs. In addition, our method provides a new and effective component to enrich the search space.

\section{Dynamic Convolutional Neural Networks}
We describe dynamic convolutional neural networks (DY-CNNs) in this section. The goal is to provide better trade-off between network performance and computational burden, within the scope of efficient neural networks. The two most popular strategies to boost the performance are making neural networks ``deeper" or ``wider". However, they both incur heavy computation cost, thus are not friendly to efficient neural networks. 

We propose dynamic convolution, which does not increase either the depth or the width of the network, but increase the model capability by aggregating multiple convolution kernels via attention. Note that these kernels are assembled differently for different input images, from where \textit{dynamic convolution} gets its name. In this section, We firstly define the generic dynamic perceptron, and then apply it to convolution.

\subsection{Preliminary: Dynamic Perceptron}
\noindent \textbf{Definition}: Let us denote the traditional or static perceptron as $\bm{y}=g(\bm{W}^T\bm{x}+\bm{b})$, where $\bm{W}$ and $\bm{b}$ are weight matrix and bias vector, and $g$ is an activation function (e.g. ReLU \cite{NairH10Relu,JarrettKRL09Relu}). We define the dynamic perceptron by aggregating multiple ($K$) linear functions $\{\bm{\tilde{W}}^T_k\bm{x}+\bm{\tilde{b}}_k\}$ as follows:
\begin{align}
\bm{y} &= g(\bm{\tilde{W}}^T(\bm{x})\bm{x}+\bm{\tilde{b}}(\bm{x})) \nonumber \\
\bm{\tilde{W}}(\bm{x})&=\sum_{k=1}^K\pi_k(\bm{x})\bm{\tilde{W}}_k, \:
\bm{\tilde{b}}(\bm{x})=\sum_{k=1}^K\pi_k(\bm{x})\bm{\tilde{b}}_k \nonumber \\
\text{s.t.} \;\;\; &0 \leq \pi_k(\bm{x}) \leq 1, 
\sum_{k=1}^K \pi_k(\bm{x}) = 1,
\label{eq:dynamic-perceptron}
\end{align}
where $\pi_k$ is the attention weight for the $k^{th}$ linear function $\bm{\tilde{W}}^T_k\bm{x}+\bm{\tilde{b}}_k$. Note that the aggregated weight $\bm{\tilde{W}}(\bm{x})$ and bias
$\bm{\tilde{b}}(\bm{x})$ are functions of input and share the same attention.

\noindent \textbf{Attention}: the attention weights $\{\pi_k(\bm{x})\}$ are not fixed, but vary for each input $\bm{x}$. They represent the optimal aggregation of linear models $\{\bm{\tilde{W}}^T_k\bm{x}+\bm{\tilde{b}}_k\}$ for a given input. 
The aggregated model $\bm{\tilde{W}}^T(\bm{x})\bm{x}+\bm{\tilde{b}}(\bm{x})$ is a non-linear function. Thus, dynamic perceptron has more representation power than its static counterpart.

\noindent \textbf{Computational Constraint}: compared with static perceptron, dynamic perceptron has the same number of output channels but bigger model size. It also introduces two additional computations: (a) computing the attention weights $\{\pi_k(\bm{x})\}$, and (b) aggregating parameters based upon attention $\sum_k\pi_k\bm{\tilde{W}}_k$ and $\sum_k\pi_k\bm{\tilde{b}}_k$. The additional computational cost should be significantly less than the cost of computing $\bm{\tilde{W}}^T\bm{x}+\bm{\tilde{b}}$. Mathematically, the computational constraint can be represented as follows:
\begin{align}
O(\bm{\tilde{W}}^T\bm{x}+\bm{\tilde{b}}) \gg O\left(\sum\pi_k\tilde{\bm{W}}_k\right)+&O\left(\sum\pi_k\tilde{\bm{b}}_k\right) \nonumber \\
+&O\left(\pi(\bm{x})\right)
\label{eq:dp-efficient}
\end{align}
where $O(\cdot)$ measures the computational cost (e.g. FLOPs). Note that fully connected layer does not satisfy this, while convolution is a proper fit for this constraint.

\subsection{Dynamic Convolution}
\begin{figure}[t]
	\begin{center}
		\includegraphics[width=0.94\linewidth]{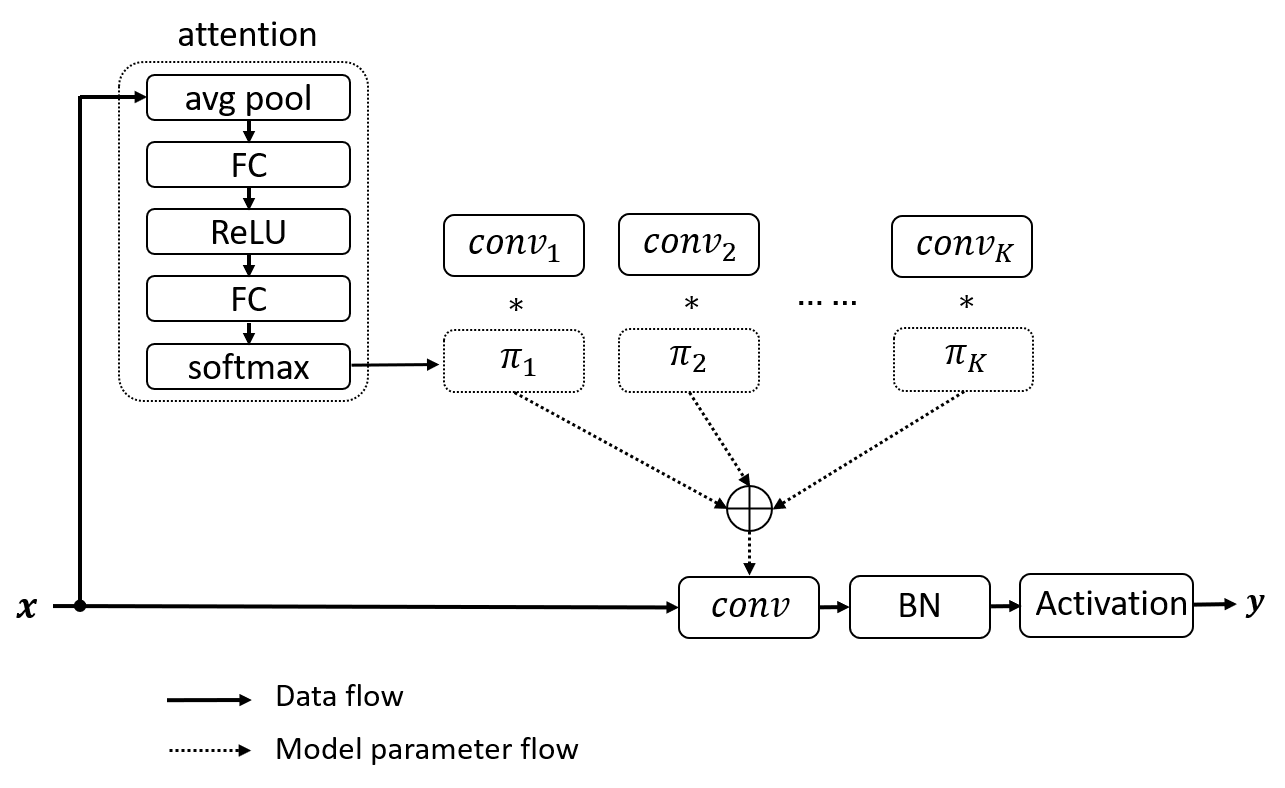}
	\end{center}
	\caption{A dynamic convolution layer.}
	\label{fig:dynamic-conv-diag}
\end{figure}

In this subsection, we showcase a specific dynamic perceptron, dynamic convolution that satisfies the computational constraint (Eq. \ref{eq:dp-efficient}). Similar to dynamic perceptron, dynamic convolution (Figure \ref{fig:dynamic-conv-diag}) has $K$ convolution kernels that share the same kernel size and input/output dimensions. They are aggregated by using the attention weights $\{\pi_k\}$. Following the classic design in CNN, we use batch normalization and an activation function (e.g. ReLU) after the aggregated convolution to build a dynamic convolution layer.

%

\noindent \textbf{Attention}: we apply squeeze-and-excitation \cite{Hu_2018_CVPR} to compute kernel attentions $\{\pi_k(\bm{x})\}$ (see Figure \ref{fig:dynamic-conv-diag}). The global spatial information is firstly squeezed by global average pooling. Then we use two fully connected layers (with a ReLU between them) and softmax to generate normalized attention weights for $K$ convolution kernels. The first fully connected layer reduces the dimension by 4. Different from SENet \cite{Hu_2018_CVPR} which computes attentions over output channels, we compute attentions over convolution kernels. The computation cost for the attention is cheap. For an input feature map with dimension $H\times W\times C_{in}$, the attention requires $O(\pi(\bm{x}))=HWC_{in}+C_{in}^2/4+C_{in}K/4$ Mult-Adds. This is much less than the computational cost of convolution, i.e. $O(\bm{\tilde{W}}^T\bm{x}+\bm{\tilde{b}})=HWC_{in}C_{out}D_k^2$ Mult-Adds, where $D_k$ is the kernel size, and $C_{out}$ is the number of output channels. 

\noindent \textbf{Kernel Aggregation:}\label{sec:complexity} aggregating convolution kernels is computationally efficient due to the small kernel size. Aggregating $K$ convolution kernels with kernel size $D_k \times D_k$, $C_{in}$ input channels and $C_{out}$ output channels introduces $KC_{in}C_{out}D_k^2+KC_{out}$ extra Multi-Adds. Compared with the computational cost of convolution ($HWC_{in}C_{out}D_k^2$), the extra cost is neligible if $K \ll HW$.
Table \ref{table:madds} shows the computational cost of using dynamic convolution in MobileNetV2. 
For instance, when using MobileNetV2 ($\times$1.0), dynamic convolution with $K=4$ kernels only increases the computation cost by 4\%. Note that even though dynamic convolution increases the model size, it does not increase the output dimension of each layer. The amount of the increase is acceptable as convolution kernels are small. 
\begin{table}[t]
	\begin{center}
		\footnotesize
		\begin{tabular}{ c|c|c| c| c}
		    \specialrule{.1em}{.05em}{.05em} 
			 & $\times1.0$ & $\times0.75$ & $\times0.5$ & $\times0.35$ \\
			\hline
			static & 300.0M & 209.0M & 97.0M & 59.2M \\ 
			\hline
			$K$=2 & 309.5M & 215.6M & 100.5M & 61.5M \\ 
			$K$=4 & 312.9M & 217.5M & 101.4M & 62.0M \\ 
			$K$=6 & 316.3M & 219.5M & 102.3M & 62.5M \\ 
			$K$=8 & 319.8M & 221.4M & 103.2M & 62.9M \\ 
			\specialrule{.1em}{.05em}{.05em}
		\end{tabular}
	\end{center}
	\caption{Mult-Adds of static convolution and dynamic convolution in MobileNetV2 with four different width multipliers ($\times1.0$, $\times0.75$, $\times0.5$, and $\times0.35$).}
	\label{table:madds}
\end{table}

\noindent \textbf{From CNNs to DY-CNNs:} dynamic convolution can be easily used as a drop-in replacement for any convolution (e.g. $1\times1$ conv, $3\times3$ conv, group convolution, depthwise convolution) in any CNN architecture. It is also complementary to other operators (like squeeze-and-excitation \cite{Hu_2018_CVPR}) and activation functions (e.g. ReLU6, h-swish \cite{howard2019mbnetv3}). In the rest of the paper, we use prefix \textbf{DY-} for the networks that use dynamic convolution. For example, DY-MobileNetV2 refers to using dynamic convolution in MobileNetV2. We also use weight $\bm{\tilde{W}}_k$ to denote a convolution kernel and ignore bias $\bm{\tilde{b}}_k$, for the sake of brevity.
\section{Two Insights of Training Deep DY-CNNs}
Training deep DY-CNNs is challenging, as it requires joint optimization of all convolution kernels $\{\bm{\tilde{W}}_k\}$ and attention model $\pi_k(\bm{x})$ across multiple layers. In this section, we discuss two insights for more \textit{efficient joint optimization}, which are crucial especially to deep DY-CNNs. 

\subsection{Insight 1: Sum the Attention to One}
\begin{figure}[t]
	\begin{center}
		\includegraphics[width=0.98\linewidth]{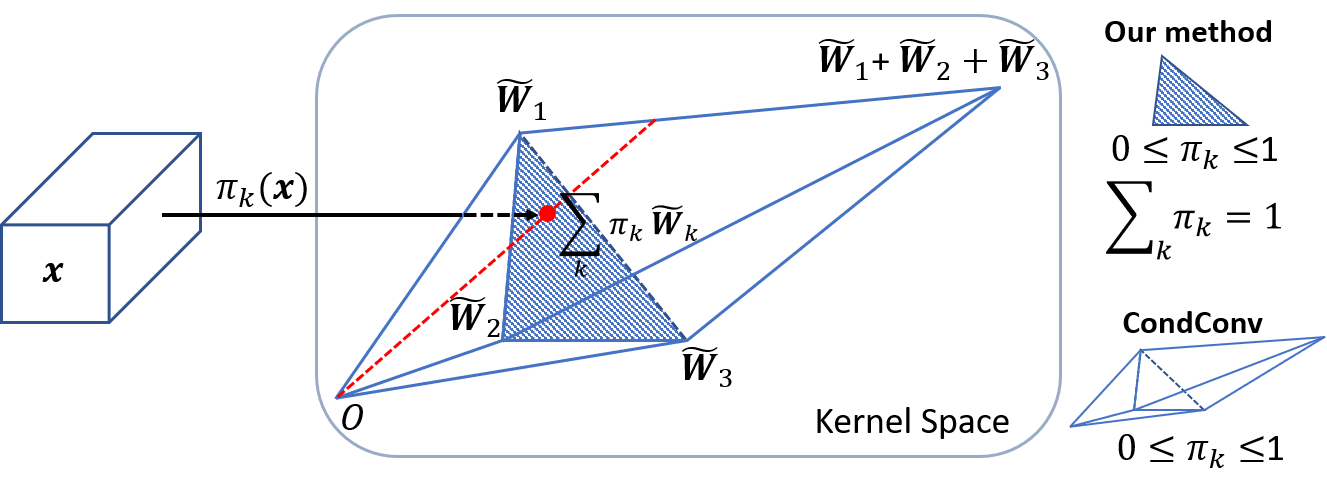}
	\end{center}
	\vspace{-4mm}
	\caption{Illustration of constraint $\sum_k \pi_k(\bm{x}) = 1$. It compresses the space of aggregated kernel $\sum_k \pi_k\bm{\tilde{W}}_k$ from two pyramids (used in CondConv \cite{Yang2019CondConvCP}) to a triangle. 
	A \textcolor{red}{red} line is compressed into a dot by normalizing attention sum. 
	Best viewed in color.}
	\label{fig:dynamic-conv}
	\vspace{-1mm}
\end{figure}

The first insight is: \textit{constraining the attention output can facilitate the learning of attention model $\pi_k(\bm{x})$.} Specifically, we have the constraint $\sum_k \pi_k(\bm{x}) = 1$ in Eq. (\ref{eq:dynamic-perceptron}) to keep the aggregated kernel $\bm{\tilde{W}}=\sum_k \pi_k\bm{\tilde{W}}_k$ within the convex hull of $\{\bm{\tilde{W}}_k\}$ in the kernel space.
Figure \ref{fig:dynamic-conv} shows an example with 3 convolution kernels. The constraint $0 \leq \pi_k(\bm{x}) \leq 1$ only keeps the aggregated kernel within the two pyramids. The sum-to-one constraint further compresses the kernel space to a triangle. It compresses the \textcolor{red}{red} line that comes from the origin into a dot by normalizing the attention sum. This normalization significantly simplifies the learning of $\pi_k(\bm{x})$, when it is jointly optimized with $\{\bm{\tilde{W}}_k\}$ in a deep network. Softmax is a natural choice of $\sum_k \pi_k(\bm{x}) = 1$.



\subsection{Insight 2: Near-uniform Attention in Early Training Epochs}
The second insight is: \textit{near-uniform attention can facilitate the learning of all kernels $\{\bm{\tilde{W}}_k\}$ during early training epochs}. 
%
This is because near-uniform attention enables more convolution kernels to be optimized simultaneously.

\begin{figure}[t]
	\begin{center}
		\includegraphics[width=1.0\linewidth]{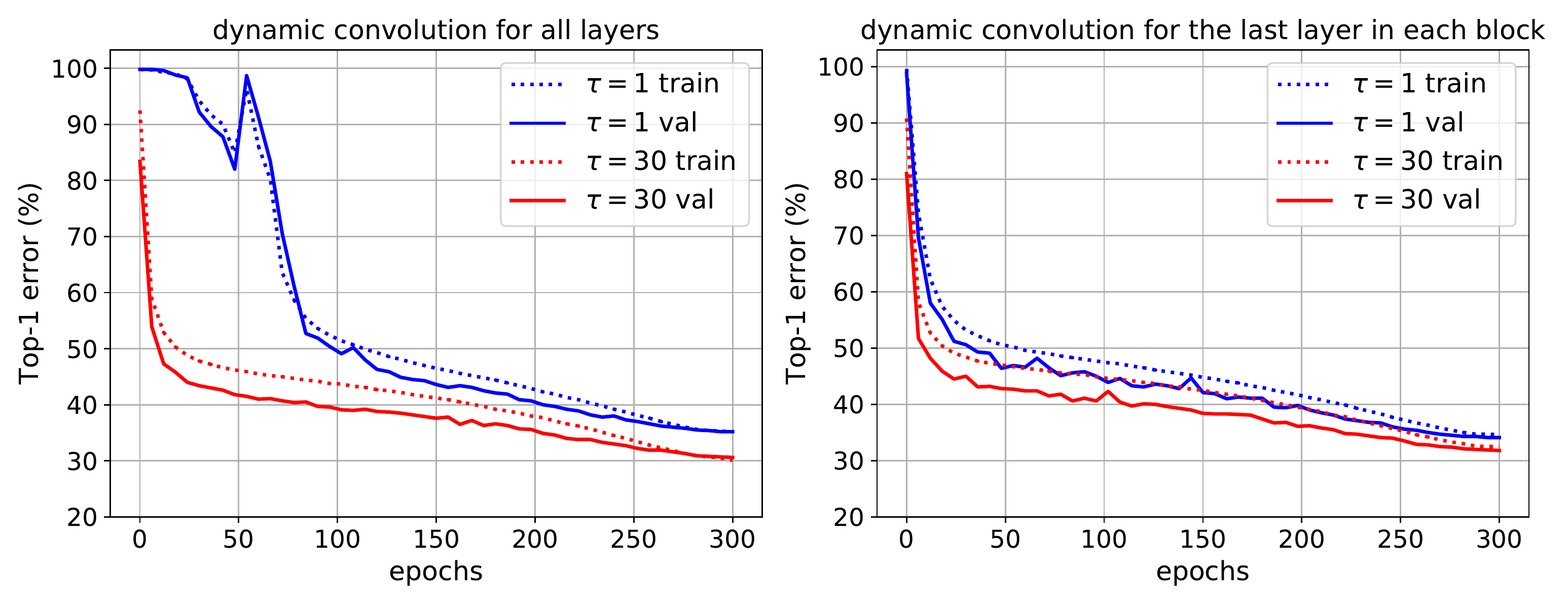}
	\end{center}
	\caption{Training and validation errors for using different softmax temperatures. \textbf{Left}: using dynamic convolution for all layers. \textbf{Right}: using dynamic convolution for the last layer in each bottleneck residual block. We use DY-MobileNetV2 with width multiplier $\times0.5$, and each dynamic convolution layer has $K=4$ convolution kernels. Best viewed in color.}
	\label{fig:tau}
\end{figure}

Softmax does NOT work well on this due to its near one-hot output. It only allows a small subset of kernels across layers to be optimized. Figure \ref{fig:tau}-(Left) shows that the training converges slowly when using softmax (\textcolor{blue}{blue} curves) to compute attention. Here, DY-MobileNetV2 with width multiplier $\times0.5$ is used. The final top-1 accuracy (64.8\%) is even worse than its static counterpart (65.4\%). 
%
This inefficiency is related to the number of dynamic convolution layers.
To validate this, we reduce the number of dynamic convolution layers by 3 (only use dynamic convolution for the last $1\times 1$ convolution in each bottleneck residual block) and expect faster convergence in training. The training and validation errors are shown in Figure \ref{fig:tau}-(Right) (\textcolor{blue}{blue} curves). As we expected, the training converges faster with higher top-1 accuracy (65.9\%) at the end.

We address this inefficiency in training deeper DY-CNNs by using a large temperature in softmax to flatten attention as follows:
\begin{align}
\pi_k=\frac{\exp(z_k/\tau)}{\sum_j\exp(z_j/\tau)},
\label{eq:gumbel}
\end{align}
where $z_k$ is the output of the second FC layer in attention branch (see Figure \ref{fig:dynamic-conv-diag}), and $\tau$ is the temperature.
The original softmax is a special case ($\tau=1$). As $\tau$ increases, the output is less sparse. When using a large temperature $\tau=30$, the training becomes significantly more efficient (see the \textcolor{red}{red} curves in Figure \ref{fig:tau}-(Left)). As a result, the top-1 accuracy boosts to 69.4\%. The larger temperature is also helpful when stacking fewer dynamic convolution layers (see \textcolor{red}{red} curves in Figure \ref{fig:tau}-(Right)). 

Temperature annealing, i.e. reducing $\tau$ from 30 to 1 linearly in the first 10 epochs, can further improve the top-1 accuracy (from 69.4\% to 69.9\%). These results support that near-uniform attention in early training epochs is crucial.


\subsection{Relation to Concurrent Work}
\begin{table}[t]
	\begin{center}
        \footnotesize
		\begin{tabular}{ c|c|c|c|c|c}
		    \specialrule{.1em}{.05em}{.05em} 
		    & Method & \#Kernels & \#Param & MAdds & Top-1   \\\hline
			\multirow{2}{*}{$\times 1.0$} &CondConv \cite{Yang2019CondConvCP} & 8 & 27.5M  & 329M & 74.6 \\
		    &DY-CNNs (ours) & 4  & 11.1M & 312.9M & \textbf{75.2}   \\
		    \hline
		    
		    \multirow{2}{*}{$\times 0.5$} & CondConv \cite{Yang2019CondConvCP} & 8 & 15.5M  & 113M & 68.4 \\
		    &DY-CNNs (ours) & 4  & 4M & 101.4M & \textbf{69.9}   \\
		    \specialrule{.1em}{.05em}{.05em}
		\end{tabular}
	\end{center}
	\caption{Comparison between DY-CNNs and the concurrent work (CondConv \cite{Yang2019CondConvCP}) on ImageNet classification using MobileNetV2 $\times 1.0$ and $\times0.5$.}
	\label{table:results}
\end{table}

These two insights are the key differences between our method and the concurrent work (CondConv \cite{Yang2019CondConvCP}), which uses sigmoid to compute kernel attention. Even though sigmoid provides near-uniform attention in early training epochs, it has significantly larger kernel space (two pyramids in Figure \ref{fig:dynamic-conv}) than our method (the shaded triangle in Figure \ref{fig:dynamic-conv}). Thus, learning attention model $\pi_k(\bm{x})$ becomes more difficult. As a result, our method has less kernels per layer, smaller model size, less computations but achieves higher accuracy (see Table \ref{table:results}). 


\section{Experiments: ImageNet Classification}
In this section, we present experimental results of dynamic convolution along with comprehensive ablations on ImageNet \cite{deng2009imagenet} classification. ImageNet has 1000 classes, including 1,281,167 images for training and 50,000 images for validation. 

\subsection{Implementation Details}
We evaluate dynamic convolution on three architectures (ResNet \cite{he2016deep}, MobileNetV2\cite{sandler2018mobilenetv2}, and MobileNetV3 \cite{howard2019mbnetv3}), by using dynamic convolution for all convolution layers except the first layer. Each layer has $K=4$ convolution kernels. The batch size is 256. We use different training setups for the three architectures as follows:

\noindent \textbf{Training setup for DY-ResNet:} The initial learning rate is 0.1 and drops by 10 at epoch 30, 60 and 90. The weight decay is 1e-4. All models are trained using SGD optimizer with 0.9 momentum for 100 epochs. We use dropout rate 0.1 before the last layer of DY-ResNet-18.

\noindent \textbf{Training setup for DY-MobileNetV2:} The initial learning rate is 0.05 and is scheduled to arrive at zero within a single cosine cycle. The weight decay is 4e-5. All models are trained using SGD optimizer with 0.9 momentum for 300 epochs. To prevent overfitting, we use label smoothing and dropout before the last layer for larger width multipliers ($\times 1.0$ and $\times 0.75$). The dropout rate are 0.2 and 0.1 for $\times 1.0$ and $\times 0.75$, respectively. 
Mixup \cite{zhang2018mixup} is used for $\times 1.0$.

\noindent \textbf{Training setup for DY-MobileNetV3:} The initial learning rate is 0.1 and is scheduled to arrive at zero within a single cosine cycle. The weight decay is 3e-5. We use SGD optimizer with 0.9 momentum for 300 epochs and dropout rate of 0.2 before the last layer. 

\subsection{Inspecting DY-CNN}
\begin{table}[t]
	\footnotesize
	\begin{center}
		\begin{tabular}{l|r r}
		    \specialrule{.1em}{.05em}{.05em} 
			Kernel Aggregation &Top-1 &	Top-5   \\
			\specialrule{.1em}{.05em}{.05em} 
			attention: $\sum\pi_k(\bm{x})\bm{\tilde{W}}_k$ & 69.4 & 88.6	\\
			\hline
			average: $\sum\bm{\tilde{W}}_k/K$  &  36.0 & 61.5  \\
			max: $\bm{\tilde{W}}_{\argmax_{k}(\pi_k)}$ & 0.1 & $0.5$	\\   
			shuffle per image: $\sum\pi_j(\bm{x})\bm{\tilde{W}}_k, j \ne k$ & 14.8 & 30.5	\\   
			shuffle across images: $\left(\sum\pi_k(\bm{x})\bm{\tilde{W}}_k\right)(\bm{x}')$ & 27.3 & 48.4	\\   
			\specialrule{.1em}{.05em}{.05em} 
		\end{tabular}
	\end{center}
	\caption{\textbf{Inspecting DY-CNN} using different kernel aggregations. DY-MobileNetV2 $\times0.5$ is used. The proper aggregation of convolution kernels $\{\bm{\tilde{W}}_k\}$ using attention $\pi_k(\bm{x})$ is shown in the first line. Shuffle per image means shuffling the attention weights for the same image over different kernels. Shuffle across images means using the attention of an image $\bm{x}$ for another image $\bm{x}'$. The poor performance for the bottom four aggregations validates that the DY-CNN is dynamic.}
	\label{table:dyanmic-or-not}
\end{table}

We inspect if DY-CNN is dynamic, using DY-MobileNetV2 $\times0.5$, which has $K=4$ kernels per layer and is trained by using $\tau=30$. Two properties are expected if it is dynamic: (a) \textit{the convolution kernels are diverse per layer}, and (b) \textit{the attention is input dependent}. 
We examine these two properties by contradiction. Firstly, if the convolution kernels are \textit{not} diverse, the performances will be stable if different attentions are used. Thus, we vary the kernel aggregation per layer in three different ways: averaging $\sum\bm{\tilde{W}}_k/K$, choosing the convolution kernel with the maximum attention $\bm{\tilde{W}}_{\argmax_{k}(\pi_k)}$, and random shuffling attention over kernels per image $\sum\pi_j(\bm{x})\bm{\tilde{W}}_k, j \ne k$.
Compared with using the original attention, the performances of these variations are significantly degraded (shown in Table \ref{table:dyanmic-or-not}). When choosing the convolution kernel with the maximum attention, the top-1 accuracy (0.1) is as low as randomly choosing a class. The significant instability confirms the diversity of convolution kernels. In addition, we shuffle attentions across images to check if the attention is input dependent. The poor accuracy (27.3\%) indicates that it is crucial for each image to use its own attention.

\begin{table}[t]
	\footnotesize
	\begin{center}
		\begin{tabular}{c c c c c|r r}
		    \specialrule{.1em}{.05em}{.05em} 
		    \multicolumn{5}{c|}{Input Resolution} & &   \\
			$112^2$ & $56^2$ & $28^2$ & $14^2$ & $7^2 $&Top-1 &	Top-5   \\
			\hline
				--	& 	--	& 	--	& 	--	& \checkmark & 57.3 & 79.9	\\
				--	& 	--	& 	--	& \checkmark & \checkmark &  67.0 & 87.2  \\
				--	& --	&\checkmark & \checkmark & \checkmark & 67.5 & 87.4	\\   
				--	&\checkmark&\checkmark & \checkmark &  \checkmark & 69.1 & 88.4	\\   
			\checkmark&\checkmark&\checkmark & \checkmark &  \checkmark & \textbf{69.4} & \textbf{88.6}	\\   
			\checkmark&\checkmark&\checkmark & \checkmark & --  & 50.9 & 76.2	\\   
			\checkmark&\checkmark&\checkmark & -- & --  & 42.5 & 68.4	\\   
			\checkmark&\checkmark&-- &--  & --  & 41.2 & 67.0	\\   
			\checkmark&--&-- & -- & --  & 37.9 & 63.5	\\   
			--&--&-- & -- & --  & 36.0 & 61.5	\\   
			\specialrule{.1em}{.05em}{.05em} 
		\end{tabular}
	\end{center}
	\caption{\textbf{Inspecting DY-CNN} by enabling/disabling attention at different input resolutions. DY-MobileNetV2 $\times0.5$ is used. 
	Each resolution has two options: 
	\checkmark indicates \textit{enabling} attention $\sum\pi_k(\bm{x})\bm{\tilde{W}}_k$ for each layer in that resolution,  
	while $-$ indicates \textit{disabling} attention and using average kernel $\sum\bm{\tilde{W}}_k/K$ for each layer in the corresponding resolutions. The attention is more effective at higher layers with lower resolutions. }
	\label{table:dyanmic-layers}
\end{table}

Furthermore, we inspect the attention across layers and find that attentions are flat at low levels and sparse at high levels. This is helpful to explain why variations in Table \ref{table:dyanmic-or-not} have poor accuracy. For instance, averaging kernels with sparse attention at high levels or picking one convolution kernel (with the maximum attention) at low levels (where attention is flat) is problematic. Table \ref{table:dyanmic-layers} shows how attention 
affects the performance across layers. We group layers by their input resolutions, and switch on/off attention for these groups. If attention is switched off for a resolution, each layer in that resolution aggregates kernels by averaging. When enabling attention at higher levels alone (resolution $14^2$ and $7^2$), the top-1 accuracy is 67.0\%, close to the performance (69.4\%) of using attention for all layers. If attention is used for lower levels alone (resolution $112^2$, $56^2$ and $28^2$), the top-1 accuracy is poor 42.5\%.

\subsection{Ablation Studies}
We perform a number of ablations on DY-MobileNetV2 and DY-MobileNetV3. The default setup includes using $K=4$ kernels per layer and $\tau=30$.
\begin{figure*}[t]
	\small
	\begin{center}
		\includegraphics[width=1.0\linewidth]{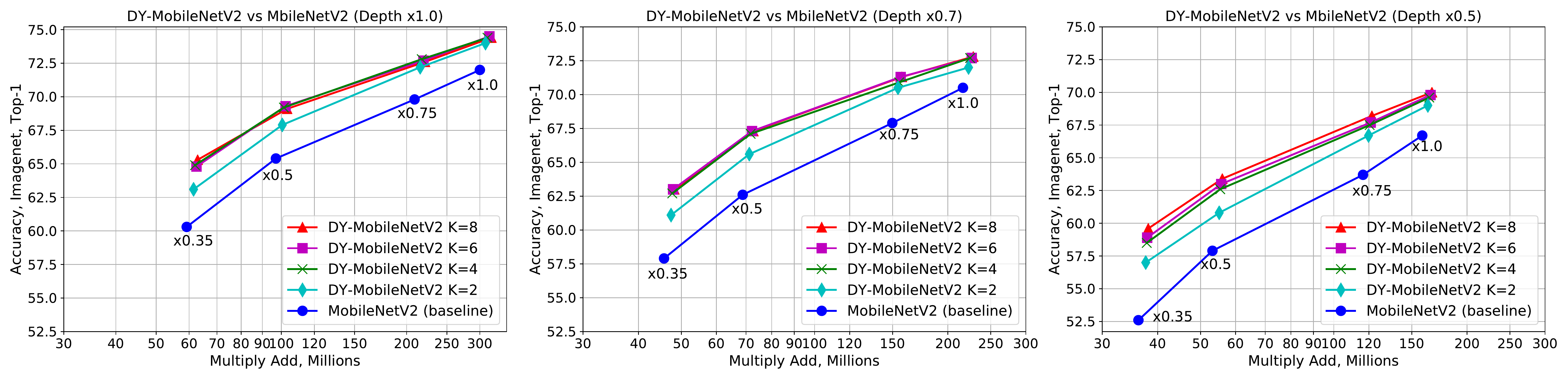}
	\end{center}
	\caption{\textbf{The number of convolution kernels} ($K$) in DY-MobileNetV2 with different depth and width multipliers. \textbf{Left}: depth multiplier is 1.0, \textbf{Middle}: depth multiplier is 0.7, \textbf{Right}: depth multiplier is 0.5. Each curve has four width multipliers $\times 1.0$, $\times 0.75$, $\times 0.5$, and $\times 0.35$. Dynamic convolution outperforms its static counterpart by a clear margin for all width/depth multipliers. Best viewed in color.}
	\label{fig:dynamic-depth-width}
\end{figure*}

\noindent \textbf{The number of convolution kernels} ($K$): the hyper-parameter $K$ controls the model complexity. Figure \ref{fig:dynamic-depth-width} shows the classification accuracy and computational cost for dynamic convolution with different $K$. We compare DY-MobileNetV2 with MobileNetV2 on different depth/width multipliers. Firstly, the dynamic convolution outperforms its static counterpart for all depth/width multipliers, even with small $K=2$. This demonstrates the strength of our method. In addition, the accuracy stops increasing once $K$ is larger than 4. This is because as $K$ increases, even though the model has more representation power, it is more difficult to optimize all convolution kernels and attention simultaneously and the network is more prone to over-fitting.

\noindent \textbf{Dynamic convolution at different layers:} Table \ref{table:dynamic-pos} shows the classification accuracy for using dynamic convolution at three different layers ($1\times1$ conv, $3\times3$ depthwise conv, $1\times1$ conv) per bottleneck residual block in MobileNetV2 $\times 0.5$. The accuracy increases as more dynamic convolution layers are used. Using dynamic convolution for all three layers yields the best accuracy. If only one layer is allowed to use dynamic convolution, using it for the last $1\times1$ convolution yields the best performance. 

\begin{table}[t]
	\begin{center}
		\footnotesize
		\begin{tabular}{r|c c c |l l}
		    \specialrule{.1em}{.05em}{.05em} 
			Network & C1 & C2 &C3 &Top-1 &	Top-5   \\
			\specialrule{.1em}{.05em}{.05em} 
			MobileNetV2  & 1 & 1 & 1 &  65.4 & 86.4  \\
			\hline
			 & 4 & 1 & 1 & 67.4$_{(2.0)}$ & 87.5$_{(1.1)}$	\\
			 & 1 & 4 & 1 & 67.4$_{(2.0)}$ & 87.3$_{(0.9)}$	\\   
			 & 1 & 1 & 4 & 68.2$_{(2.8)}$ & 87.9$_{(1.5)}$	\\      
			DY-MobileNetV2 & 4 & 1 & 4 & 68.7$_{(3.3)}$ & 88.0$_{(1.6)}$	\\   
			 & 1 & 4 & 4 & 68.4$_{(3.0)}$ & 87.9$_{(1.5)}$	\\   
			 & 4 & 4 & 1 & 68.6$_{(3.2)}$ & 88.0$_{(1.6)}$	\\   
			 & 4 & 4 & 4 & \textbf{69.4}$\bm{_{(4.0)}}$ & \textbf{88.6}$\bm{_{(2.2)}}$	\\   
			\specialrule{.1em}{.05em}{.05em} 
		\end{tabular}
	\end{center}
	\caption{\textbf{Dynamic convolution at different layers} in MobileNetV2 $\times 0.5$. C1, C2 and C3 indicate the $1\times1$ convolution that expands output channels, the $3\times3$ depthwise convolution and the $1\times1$ convolution that shrinks output channels per block respectively. C1=1 indicates using static convolution, while C1=4 indicates using dynamic convolution with 4 kernels. The numbers in brackets denote the improvement over the baseline.}
	\label{table:dynamic-pos}
\end{table}

\noindent \textbf{Softmax Temperature}: the temperature $\tau$ in softmax controls the sparsity of attention weights. It is important for training DY-CNNs effectively. Table \ref{table:dyanmic-tau} shows the classification accuracy for using different temperatures. $\tau=30$ has the best performance. Furthermore, temperature annealing (reducing $\tau$ from 30 to 1 linearly in the first 10 epochs) provides additional improvement on top-1 accuracy (from 69.4\% to 69.9\%). Therefore, using large temperature in the early stage of training is important.

\begin{table}[t]
	\begin{center}
		\footnotesize
		\begin{tabular}{r|l|l l}
		    \specialrule{.1em}{.05em}{.05em} 
			Network & Temperature &Top-1 &	Top-5   \\
			\specialrule{.1em}{.05em}{.05em} 
			MobileNetV2  & \multicolumn{1}{c|}{---} &  65.4 & 86.4  \\
			\hline
			 & $\tau=1$ & 64.8$_{(-0.6)}$ & 85.5$_{(-0.9)}$	\\   
			 & $\tau=5$ & 65.7$_{(+0.3)}$ & 85.8$_{(-0.6)}$	\\   
			 & $\tau=10$ & 67.5$_{(+2.1)}$ & 87.4$_{(+1.0)}$	\\   
			DY-MobileNetV2 & $\tau=20$ & \textbf{69.4}$\bm{_{(+4.0)}}$ & 88.5${_{(+2.1)}}$	\\   
			 & $\tau=30$ & \textbf{69.4}$\bm{_{(+4.0)}}$ & \textbf{88.6}$\bm{_{(+2.2)}}$	\\   
			 & $\tau=40$ & 69.2${_{(+3.8)}}$ & 88.4${_{(+2.0)}}$	\\   
			\cline{2-4}
			 & $\tau$ annealing & \textbf{69.9}$_{\bm{(+4.5)}}$ & \textbf{89.0}$_{\bm{(+2.6)}}$  \\
 			\specialrule{.1em}{.05em}{.05em} 
		\end{tabular}
	\end{center}
	\caption{\textbf{Softmax Temperature}: large temperature in early training epochs is important. Temperature annealing refers to reducing $\tau$ from 30 to 1 linearly in the first 10 epochs. The numbers in brackets denote the performance improvement over the baseline.}
	\label{table:dyanmic-tau}
\end{table}

\noindent \textbf{Dynamic Convolution vs Squeeze-and-Excitation (SE) \cite{Hu_2018_CVPR}}: MobileNetV3-Small \cite{howard2019mbnetv3} is used, in which the locations of SE layers are considered optimal as they are found by network architecture search (NAS). The results are shown in Table \ref{table:dynamic-se}. Without using SE, the top-1 accuracy for MobileNetV3-Small drops 2\%. However, DY-MobileNetV3-Small \textit{without} SE outperforms MobileNetV3-Small with SE by 2.2\% in top-1 accuracy. Combining dynamic convolution and SE gains additional 0.7\% improvement. This suggests that attention over kernels and attention over output channels can work together.

\begin{table}[t]
	\begin{center}
		\footnotesize
		\begin{tabular}{l|l l}
		    \specialrule{.1em}{.05em}{.05em} 
			Network &Top-1 &	Top-5   \\
			\specialrule{.1em}{.05em}{.05em} 
			MobileNetV3-Small  &   67.4 & 86.4   \\
			MobileNetV3-Small w/o SE  & 65.4$_{(-2.0)}$ & 85.2$_{(-1.2)}$	 		 \\   
			\hline
			DY-MobileNetV3-Small  &   70.3$_{(+2.9)}$ & 88.7$_{(+2.3)}$	 		 \\
			Dy-MobileNetV3-Small w/o SE  &   69.6$_{(+2.2)}$ & 88.4$_{(+2.0)}$	 		 \\
			\specialrule{.1em}{.05em}{.05em} 
		\end{tabular}
	\end{center}
	\caption{\textbf{Dynamic convolution vs Squeeze-and-Excitation (SE \cite{Hu_2018_CVPR})} on MobileNetV3-Small. The numbers in brackets denote the performance improvement over the baseline. Compared with static convolution with SE, dynamic convolution \textit{without} SE gains 2.2\% top-1 accuracy.}
	\label{table:dynamic-se}
\end{table}

\subsection{Main Results}

Table \ref{table:imagenet-cls-result} shows the comparison between dynamic convolution and its static counterpart in three CNN architectures (MobileNetV2, MobileNetV3 and ResNet). $K=4$ kernels are used in each dynamic convolution layer and temperature annealing is used in the training. Although we focus on efficient CNNs, we evaluate dynamic convolution on two shallow ResNets (ResNet-10 and ResNet-18) to show its effectiveness on $3\times3$ convolution, which is only used for the first layer in MobileNet V2 and V3. Without bells and whistles, dynamic convolution outperforms its static counterpart by a clear margin for all three architectures, with small extra computational cost ($\sim4\%$). DY-ResNet and DY-MobileNetV2 gains more than 2.3\% and 3.2\% top-1 accuracy, respectively. DY-MobileNetV3-Small is 2.9\% more accurate than the state-of-the-art MobileNetV3-Small. 



\begin{table}[t!]
	\begin{center}
	    \scriptsize
		\begin{tabular}{r|r|r|ll}
		    \specialrule{.1em}{.05em}{.05em} 
			Network & \#Param & MAdds &Top-1 &	Top-5   \\
		
			\specialrule{.1em}{.05em}{.05em} 
			MobileNetV2 $\times1.0$  & 3.5M & 300.0M & 72.0  & 91.0  \\
			DY-MobileNetV2 $\times1.0$ & 11.1M & 312.9M & 75.2$_{(3.2)}$ & 92.1$_{(1.1)}$	  \\
			 \hline
			MobileNetV2 $\times0.75$  & 2.6M & 209.0M & 69.8 & 89.6 		 \\
			DY-MobileNetV2 $\times0.75$ & 7.0M & 217.5M  & 73.7$_{(3.9)}$ & 91.3$_{(1.7)}$		 \\
			\hline
			MobileNetV2 $\times0.5$  & 2.0M & 97.0M & 65.4 & 86.4 		 \\
			DY-MobileNetV2 $\times0.5$ & 4.0M & 101.4M & 69.9$_{(4.5)}$ & 89.0$_{(2.6)}$		 \\
			\hline
			MobileNetV2 $\times0.35$ & 1.7M & 59.2M  & {60.3}& {82.9} 		 \\
			DY-MobileNetV2 $\times0.35$& 2.8M & 62.0M & 65.9$_{(5.6)}$ & 86.4$_{(3.5)}$ 		 \\
			\hline
			MobileNetV3-Small  & 2.9M & 66.0M & 67.4 & 86.4   \\
			DY-MobileNetV3-Small  & 4.8M & 68.5M & 70.3$_{(2.9)}$ & 88.7$_{(2.3)}$	 		 \\
			\hline
			ResNet-18  & 11.1M & 1.81G & 70.4  & 89.7  \\
			DY-ResNet-18 & 42.7M & 1.85G & 72.7$_{(2.3)}$ & 90.7$_{(1.0)}$	  \\
			\hline
			ResNet-10 & 5.2M & 0.89G & 63.5  & 85.0  \\
			DY-ResNet-10 & 18.6M & 0.91G & 67.7$_{(4.2)}$ & 87.6$_{(2.6)}$	  \\
			\specialrule{.1em}{.05em}{.05em} 
		\end{tabular}
	\end{center}
	\caption{ImageNet \cite{deng2009imagenet} classification results of DY-CNNs.
	The numbers in brackets denote the performance improvement over the baseline.}
	\label{table:imagenet-cls-result}
\end{table}

\section{DY-CNNs for Human Pose Estimation}
We use COCO 2017 dataset \cite{lin2014microsoft} to evaluate dynamic convolution on single-person keypoint detection. Our models are trained on \texttt{train2017}, including $57K$ images and $150K$ person instances labeled with 17 key-points. We evaluate our method on \texttt{val2017} containing 5000 images and use the mean average precision (AP) over 10 object key point similarity (OKS) thresholds as the metric.

\begin{table}[t]
	\begin{center}
		\footnotesize
		\begin{tabular}{c|c|c|c|c}
		    \specialrule{.1em}{.05em}{.05em} 
			Input & Operator &	exp size & $\#out$ & $n$   \\
			\hline
			$16\times12\times B_{out}$  & bneck, $5\times5$ & 768 & 256 & 2  \\
			$32\times24\times 256$  & bneck, $5\times5$ & 768 & 128 & 1  \\   
			$64\times48\times 128$  & bneck, $5\times5$ & 384 & 128 & 1  \\
			\specialrule{.1em}{.05em}{.05em} 
		\end{tabular}
	\end{center}
	\caption{Light-weight head structures for keypoint detection. We use MobileNetV2's bottleneck residual block \cite{sandler2018mobilenetv2} (denoted as bneck). Each row is corresponding to a stage, which starts with a bilinear upsampling operator to scale up the feature map by 2. $\#out$ denotes the number of output channels, and $n$ denotes the number of bottleneck residual blocks.}
	\label{table:pose-head}
\end{table}

\noindent \textbf{Implementation Details:} We implement two types of networks to evaluate dynamic convolution. \textit{Type-A} follows SimpleBaseline \cite{xiao2018simplebaseline} by using deconvolution in head. We use MobileNetV2 and V3 as a drop-in replacement for the backbone feature extractor and compare static convolution and dynamic convolution in the \textit{backbone alone}. \textit{Type-B} still uses MobileNetV2 and V3 as backbone. But it uses upsampling and MobileNetV2's bottleneck residual block in head. We compare dynamic convolution with its static counterpart in \textit{both backbone and head}. The details of head structure are shown in Table \ref{table:pose-head}. For both types, we use $K=4$ kernels in each dynamic convolution layer.

\noindent \textbf{Training setup:} We follow the training setup in \cite{sun2019deep}. The human detection boxes are cropped from the image and resized to $256\times192$. The data augmentation includes random rotation ($[-\ang{45}, \ang{45}]$), random scale ($[0.65, 1.35]$), flipping, and half body data augmentation. All models are trained from scratch for 210 epochs, using Adam optimizer \cite{kingma:adam}. The initial learning rate is set as 1e-3 and is dropped to 1e-4 and 1e-5 at the $170^{th}$ and $200^{th}$ epoch, respectively. The temperature of softmax in DY-CNNs is set as $\tau=30$.

\noindent \textbf{Testing:} We follow \cite{xiao2018simplebaseline, sun2019deep} to use two-stage top-down paradigm: detecting person instances using a person detector and then predicting keypoints. We use the same person detectors provided by \cite{xiao2018simplebaseline}. The keypoints are predicted on the average heatmap of the original and flipped images by adjusting the highest heat value location with a quarter offset from the highest response to the second highest response.

\begin{table*}[t]
	\begin{center}
		\footnotesize
		\begin{tabular}{c|r c r| c c r|l c c c c c}
			\specialrule{.1em}{.05em}{.05em} 
			\multirow{2}{*}{Type} &\multicolumn{3}{c|}{Backbone}   &	\multicolumn{3}{c|}{Head} & &     &   &   & & \\
			                      & Networks&  \#Param & MAdds & Operator & \#Param & MAdds  & AP &	AP$^{0.5}$ & AP$^{0.75}$ & AP$^M$ & AP$^L$&	AR  \\
			\specialrule{.1em}{.05em}{.05em} 
			\multirow{2}{*}{A}&ResNet-18 & 10.6M & 1.77G & dconv & 8.4M & 5.4G     & 67.0 & 87.9 & 74.8 & 63.6 & 73.5 & 73.1 \\
			&DY-ResNet-18& 42.2M & 1.81G & dconv & 8.4M & 5.4G    & 68.6$_{(1.6)}$ &	88.4 & 76.1 & 65.3 & 75.1 & 74.6 \\
			\hline
			\multirow{2}{*}{A}&MobileNetV2 $\times1.0$  & 2.2M & 292.6M & dconv & 8.4M & 5.4G   &  64.7         & 87.2 & 72.6 & 61.3 & 71.0 & 71.0 \\
			&DY-MobileNetV2 $\times1.0$& 9.8M & 305.3M & dconv & 8.4M & 5.4G  &67.6$_{(2.9)}$ &	88.1 & 75.5 & 64.4 & 74.1 & 73.8 \\
			\hline
			\multirow{2}{*}{A}&MobileNetV2 $\times0.5$  & 0.7M & 93.7M & dconv & 8.4M & 5.4G   &{57.0}         & {83.7} & 	{63.1} &	{53.9}&		{63.1}&	{63.7}	 \\
			&DY-MobileNetV2 $\times0.5$& 2.7M & 98.0M & dconv & 8.4M & 5.4G  &61.9$_{(4.9)}$ & 	85.8 &	69.7&		58.9&	67.9&	68.4	 \\
			
			\hline
			\multirow{2}{*}{A}&MobileNetV3-Small        & 1.1M & 62.7M & dconv & 8.4M & 5.4G   &{57.1}& {83.7} & 	{63.8} &	{54.9}&		{62.3}&	{64.1}	 \\
			&DY-MobileNetV3-Small& 2.8M & 65.1M & dconv & 8.4M & 5.4G  &59.3$_{(2.2)}$ & 84.7	 &	66.7 &	56.9	&	64.7&	66.1	 \\
			\specialrule{.1em}{.05em}{.05em} 
			\multirow{2}{*}{B}&MobileNetV2 $\times1.0$  & 2.2M & 292.6M & bneck & 1.2M & 701.1M  &64.6           &87.0  & 72.4 & 61.3 & 71.0 & 71.0 \\
			&DY-MobileNetV2 $\times1.0$& 9.8M & 305.3M & bneck & 6.3M & 709.4M &68.2$_{(3.6)}$ &	88.4 & 76.0 & 65.0 & 74.7 & 74.2 \\
			\hline
			\multirow{2}{*}{B}&MobileNetV2 $\times0.5$  & 0.7M & 93.7M & bneck & 1.2M & 701.1M   &{59.2}         & {84.3} & 	{66.4} &	{56.2}&		{65.0}&	{65.6}	 \\
			&DY-MobileNetV2 $\times0.5$& 2.7M & 98.0M & bneck & 6.3M& 709.4M &62.8$_{(3.6)}$ & 	86.1 &	70.4&		59.9&	68.6&	69.1	 \\
			
			\hline
			\multirow{2}{*}{B}&MobileNetV3-Small        & 1.1M & 62.7M & bneck & 1.0M & 664.2M   & 57.1 & 83.8 & 63.7 &	55.0&		62.2&	64.1	 \\
			&DY-MobileNetV3-Small& 2.8M & 65.1M & bneck & 4.9M & 671.1M  &60.0$_{(2.9)}$ & 85.0	 &	67.8 &	57.6	&	65.4&	66.7	 \\		
			\specialrule{.1em}{.05em}{.05em} 
		\end{tabular}
	\end{center}
	\caption{Keypoint detection results on COCO validation set. All models are trained from scratch. The top half uses dynamic convolution in the backbone and uses deconvolution in the head (Type A). The bottom half use MobileNetV2's bottleneck residual blocks in the head and use dynamic convolution in both the backbone and the head (Type B). Each dynamic convolution layer includes $K=4$ kernels. The numbers in brackets denote the performance improvement over the baseline.}
	\label{table:coco-kp}
\end{table*}

\noindent \textbf{Main Results and Ablations:} Firstly we compare dynamic convolution with its static counterpart in the backbone (\textit{Type-A}). The results are shown in the top half of Table \ref{table:coco-kp}. Dynamic convolution gains 1.6, 2.9, 2.2 AP for ResNet-18, MobileNetV2 and MobileNetV3-Small, respectively. 

Secondly, we replace the heavy deconvolution head with light-weight upsampling and MobileNetV2's bottleneck residual blocks (\textit{Type-B}) to make the whole network small and efficient. Thus, we can compare dynamic convolution with its static counterpart in both backbone and head. The results are shown in the bottom half of Table \ref{table:coco-kp}. Similar to \textit{Type-A}, dynamic convolution outperforms its static counterpart by a clear margin. It gains 3.6 and 2.9 AP for MobileNetV2 and MobileNetV3-Small, respectively. 

We perform an ablation to investigate the effects of dynamic convolution at backbone and head separately (Table \ref{table:pose-ablation}). Even though most of improvement comes from the dynamic convolution at the backbone, dynamic convolution at the head is also helpful. This is mainly because the backbone has more convolution layers than the head.

\begin{table}[t]
	\footnotesize
	\begin{center}
		\begin{tabular}{c|c|l c c}
		    \specialrule{.1em}{.05em}{.05em} 
			Backbone & Head & AP & AP$^{0.5}$ & AP$^{0.75}$  \\
			\hline
			static      & static    & 59.2          & 84.3 & 66.4  \\
			static      & dynamic   & 60.3$_{(1.1)}$ & 84.9 & 67.3  \\   
			dynamic     & static    & 62.3$_{(3.1)}$  & 85.6 & 70.0  \\
			dynamic     & dynamic   & 62.8$_{(3.6)}$ & 86.1 & 70.4  \\
			\specialrule{.1em}{.05em}{.05em} 
		\end{tabular}
	\end{center}
	\caption{Keypoint detection results of using dynamic convolution in backbone and head separately. We use MobileNetV2 $\times0.5$ as backbone and use the light-weight head structure discussed in Table \ref{table:pose-head}. The numbers in brackets denote the performance improvement over the baseline. Dynamic convolution can improve AP at both the backbone and the head.}
	\label{table:pose-ablation}
\end{table}

\section{Conclusion}
In this paper, we introduce dynamic convolution, which aggregates multiple convolution kernels dynamically based upon their attentions for each input. Compared to its static counterpart (single convolution kernel per layer), it significantly improves the representation capability with negligible extra computation cost, thus is more friendly to efficient CNNs. Our dynamic convolution can be easily integrated into existing CNN architectures. By simply replacing each convolution kernel in MobileNet (V2 and V3) with dynamic convolution, we achieve solid improvement for both image classification and human pose estimation. We hope dynamic convolution becomes a useful component for efficient network architectures.

\clearpage
\appendix
\section{Appendix}

In this appendix, we report running time and perform additional analysis for our dynamic convolution method.

\subsection{Inference Running Time}
We report the running time of dynamic MobileNetV2 (DY-MobileNetV2) with four different width multipliers ($\times1.0$, $\times0.75$, $\times0.5$, and $\times0.35$) and compare with its static counterpart (MobileNetV2 \cite{sandler2018mobilenetv2}) in Table \ref{table:imagenet-runtime}. We use a single-threaded core of Intel Xeon CPU E5-2650 v3 (2.30GHz) to measure running time (in milliseconds). The running time is calculated by averaging the inference time of 5,000 images with batch size 1. Both MobileNetV2 and DY-MobileNetV2 are implemented using PyTorch \cite{paszke2017automatic}.

Compared with its static counterpart, DY-MobileNetV2 consumes about $10\%$ more running time and $4\%$ more Multi-Adds. The overhead of running time is higher than Multi-Adds. We believe this is because the optimizations of global average pooling and small inner-product operations are not as efficient as convolution. With the small additional computational cost, our dynamic convolution method significantly improves the model performance.

\begin{table}[b]
	\begin{center}
	    \footnotesize
		\begin{tabular}{r|l| r|c }
		    \specialrule{.1em}{.05em}{.05em} 
			Network &Top-1 & MAdds	& CPU (ms)   \\
		
			\hline
			MobileNetV2 $\times1.0$  & 72.0 & 300.0M & 127.9  \\
			DY-MobileNetV2 $\times1.0$ & 75.2$_{(3.2)}$ & 312.9M  & 141.2	  \\
			 \hline
			MobileNetV2 $\times0.75$  & 69.8 & 209.0M  & 99.5 \\
			DY-MobileNetV2 $\times0.75$ & 73.7$_{(3.9)}$ & 217.5M   & 110.5		 \\
			\hline
			MobileNetV2 $\times0.5$  & 65.4 &  97.0M  & 69.6 		 \\
			DY-MobileNetV2 $\times0.5$ & 69.9$_{(4.5)}$ & 101.4M  & 77.4		 \\
			\hline
			MobileNetV2 $\times0.35$ & {60.3} & 59.2M & 61.1 		 \\
			DY-MobileNetV2 $\times0.35$ & 65.9$_{(5.6)}$ & 62.0M  &  67.4		 \\
			
			\specialrule{.1em}{.05em}{.05em} 
		\end{tabular}
	\end{center}
	\caption{\textbf{Inference running time} of DY-MobileNetV2 \cite{sandler2018mobilenetv2} on ImageNet \cite{deng2009imagenet} classification. We use dynamic convolution with $K=4$ kernels for all convolution layers in DY-MobileNetV2 except the first layer. CPU: CPU time in milliseconds measured on a single core of Intel Xeon CPU E5-2650 v3 (2.30GHz). The running time is calculated by averaging the inference time of 5,000 images with batch size 1. The numbers in brackets denote the performance improvement over the baseline.}
	\label{table:imagenet-runtime}
\end{table}

\subsection{Dynamic Convolution in Shallower and Thinner Networks}
\begin{figure}[b]
	\small
	\begin{center}
		\includegraphics[width=0.70\linewidth]{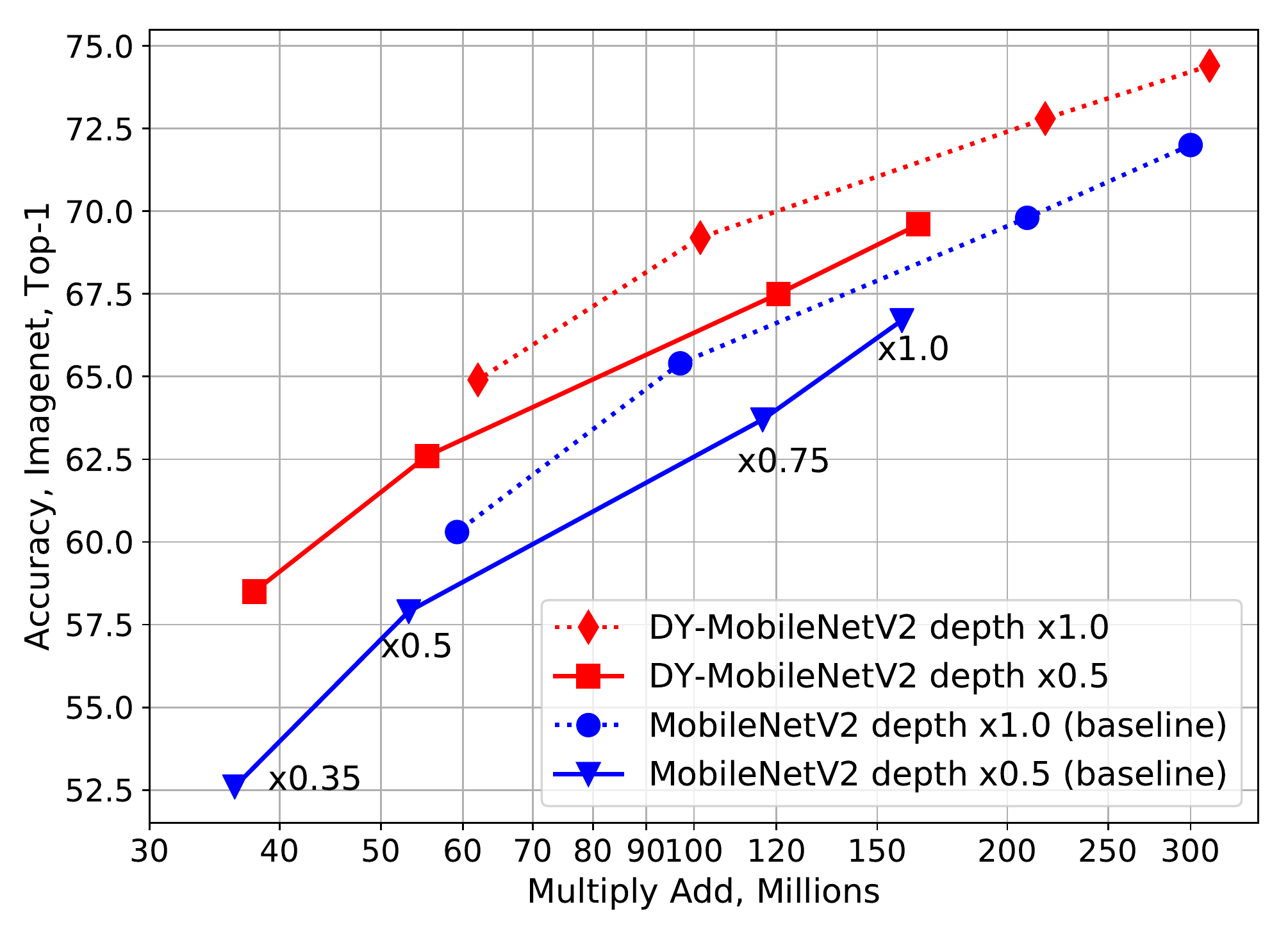}
	\end{center}
	\caption{\textbf{Shallower DY-MobileNetV2 vs Deeper MobileNetV2}. The shallower DY-MobileNetV2 (depth $\times0.5$) has better trade-off between accuracy and computational cost than the deeper MobileNetV2 (depth $\times 1.0$). To make comparison fair, we also plot the deeper DY-MobileNetV2 and shallower MobileNetV2. For both DY-MobileNetV2 and MobileNetV2, deeper networks have better performance. Best viewed in color.}
	\label{fig:dynamic-depth}
\end{figure}

 Figure \ref{fig:dynamic-depth} shows that the shallower DY-MobileNetV2 (depth $\times0.5$) has better trade-off between accuracy and computational cost than the deeper MobileNetV2 (depth $\times 1.0$), even though shallower networks (depth $\times$0.5) have performance degradation for both DY-MobileNetV2 and MobileNetV2. Improvement on shallow networks is useful as they are friendly to parallel computation.
Furthermore, dynamic convolution achieves more improvement for thinner and shallower networks with small width/depth multipliers. This is because thinner and shallower networks are underfitted due to their limited model size and dynamic convolution significantly improves their capability. 
\

\subsection{Example: Learning XOR} 
To make the idea of dynamic perceptron more concrete, we use it on a simple task, i.e. learning the XOR function. In this example, we want our network to perform correctly on the four points $\mathbb{X}=\{[0,0]^T, [0,1]^T, [1,0]^T, [1,1]^T\}$. Compared with the solution using \textit{two} static perceptron layers \cite{Goodfellow:2016:DL:3086952} as follows:
\begin{align}
\bm{y} &= \bm{w}^T\max\{0, \bm{W}^T\bm{x}+\bm{b}\} \nonumber \\
\bm{w}&=\begin{bmatrix}
    1 \\ -2
\end{bmatrix},
\bm{W}=\begin{bmatrix}
    1 & 1 \\ 1 & 1
\end{bmatrix}
\bm{b}=\begin{bmatrix}
    0 \\ -1
\end{bmatrix},
\label{eq:static-xor}
\end{align}
dynamic perception only needs a \textit{single} layer as follows:
\begin{align}
\bm{y} &= \sum_{k=1}^2\left[\left(\pi_k(\bm{x})\bm{\tilde{W}}_k^T\right)\bm{x}+\pi_k(\bm{x})\bm{\tilde{b}}_k\right] \nonumber \\
\bm{\tilde{W}}_1&=\begin{bmatrix}
    -1 & 0 \\ 0 & 0
\end{bmatrix},
\bm{\tilde{b}}_1=\begin{bmatrix}
    1 \\ 0
\end{bmatrix},
\bm{\tilde{W}}_2=\begin{bmatrix}
    1 & 0 \\ 0 & 0
\end{bmatrix},
\bm{\tilde{b}}_2=\begin{bmatrix}
    0 \\ 0
\end{bmatrix},
\label{eq:dynamic-xor}
\end{align}
where the attentions are $\pi_1(\bm{x})=x_2$, $\pi_2(\bm{x})=1-x_2$. This example demonstrates that dynamic perceptron has more representation power due to the non-linearity.  

\clearpage

{\small
\bibliographystyle{ieee_fullname}
\bibliography{egbib}
}

\end{document}